\documentclass{esannV2}
\usepackage{graphicx}
\usepackage[latin1]{inputenc}
\usepackage{amssymb,amsmath,array}
\usepackage{hyperref}
\usepackage{multirow}

\usepackage{tabularx}

\begin{document}
\title{Automatic Trade-off Adaptation in Offline RL}

\author{Phillip Swazinna$^{1,2}$, Steffen Udluft$^1$, and Thomas Runkler$^{1,2}$
%
\vspace{.3cm}\\
%
1) Siemens Technology - Data Analytics \& Artificial Intelligence \\
Otto-Hahn-Ring 6, Munich, Germany
%
\vspace{.1cm}\\
2) TU Munich - School of Computation, Information, and Technology \\
Boltzmannstrasse 3, Garching, Germany\\
}

\maketitle

\begin{abstract}
Recently, offline RL algorithms have been proposed that remain adaptive at runtime. For example, the LION algorithm \cite{lion} provides the user with an interface to set the trade-off between behavior cloning and optimality w.r.t. the estimated return at runtime. Experts can then use this interface to adapt the policy behavior according to their preferences and find a good trade-off between conservatism and performance optimization. Since expert time is precious, we extend the methodology with an autopilot that automatically finds the correct parameterization of the trade-off, yielding a new algorithm which we term AutoLION.
\end{abstract}

\section{Introduction \& Related Work}

Over the past few years, offline RL has become a popular field of reinforcement learning research, since it promises to alleviate one of the most pressing issues when trying to apply RL methods to real-world (i.e. potentially physical) systems: Online environment interaction. Direct interaction is often prohibited in real systems, since as opposed to e.g. simulated Atari video games, they incur significant (opportunity) costs and due to potential safety violations. Offline RL methods such as \cite{rambo,moose} thus constitute a large step towards broader applicability of reinforcement learning methodologies in practice, since they demonstrate the ability to learn purely from static, pre-collected datasets.\\

A remaining issue is, that each of the proposed algorithms in one way or the other takes a trade-off between optimizing for return only (i.e. pure RL) and regularizing the policy towards the dataset distribution, since otherwise trained policies would exploit the return estimating models and transfer badly to the real system. The problem is, that nobody can with certainty determine the amount of regularization that is correct, i.e. how strictly does the policy need to be regularized given the dataset remains an open question (depending on the concrete form of this regularization, different terms such as conservatism, pessimism, risk-avoidance, uncertainty-avoidance, reconstruction penalty, behavior constraint, proximity, etc. have been proposed).\\

Recently, \cite{offline} proposed that offline RL policies should be trained to be adaptive at runtime, i.e. after training has conceded. The authors propose to maintain a distribution of possible MDPs, which can then at runtime be narrowed down to perform actions that are estimated to be optimal in the currently believed MDP. \cite{lion,confidence} take this a step further and train policies that can remain adaptive in their level of regularization after training has conceded. While \cite{confidence} is developed for discrete action spaces, such as Atari video games, LION \cite{lion} is designed for continuous control environments such as the industrial benchmark (IB) \cite{IB} or MuJoCo robotics locomotion tasks. LION enables human expert users to make the concrete trade-off conditioning hyperparameter choice in order to provide them with a utility instead of completely automating the regularization trade-off as in \cite{confidence}. While this is valuable feature in practice, some users would likely welcome the option to delegate this task in some situations, such as when they are supervising many systems simultaneously, allowing them to better focus on the problematic cases. Our goal is therefore to extend LION with an autopilot mode that automatically chooses the trade-off during runtime.

\section{Preliminaries: LION}
The LION (Learning in Interactive Offline eNvironments) algorithm was proposed in \cite{lion}, introducing a model-based offline RL method to train policies that remain trade-off adaptive after training has conceded. Instead of training policies for a fixed trade-off between performance and proximity to the behavioral, it samples the trade-off parameter randomly for each starting state of a trajectory during training and conditions the policy on it, enabling policies to learn the entire range from pure behavior cloning over regularized RL up to ``pure'' (unregularized) RL.\\

The algorithm trains an ensemble of $K$ (potentially recurrent) transition models $\hat{T}^k(\cdot|s,a),\ k=0,...,K-1$ that predict the future states and rewards in an environment based on past states and actions. It then performs ``imagined'' trajectory rollouts with the target policy through the trained transition models as environment surrogates and optimizes the resulting return estimate by backpropagation through time. At the same time, the target policy actions in these trajectories are compared with the actions of a learned model of the behavioral policy $\beta(\cdot|s)$, in order to regularize the policy towards the known state-action space. The trade-off between the two components is in this case determined by the hyperparameter $\lambda$, which the policy $\pi_\theta(\cdot|s,\lambda)$ also conditions on:
\begin{equation}
    L(\theta) = \mathbb{E}_{s_0 \sim \mathcal{D},\ \lambda \sim B(a,b)} \sum_{t=0}^H \gamma^t \left[ \lambda e(s_t,\pi_\theta(s_t,\lambda)) - (1-\lambda) p(\beta(s_t),\pi_\theta(s_t,\lambda)) \right]
\end{equation}
where $\mathcal{D}$ denotes the initial dataset, $B(a,b)$ is the Beta distribution from which $\lambda$ is sampled, $e(s,a)=\min_k r(\hat{T}^k(s,a))$ denotes the minimum reward prediction by any of the ensemble members, and $p(a_\beta, a_\pi)=\mathrm{mean}(a_\beta-a_\pi)^2$ is the penalty term to regularize the policy.

\section{Automatic trade-off search with LION (AutoLION)}
The LION methodology has been proposed in order to provide expert practitioners with a utility: Instead of automating them away, users can benefit from a still highly autonomous system, yet also have the possibility to interact with the policy and alter its behavior on a higher level of abstraction, by handpicking $\lambda$ at runtime, depending on their observations and knowledge about the system. However, we find that practitioners would still benefit from a solution that integrates a more autopilot-like behavior - since expert time is always costly and limited, and since users likely look after more than just a single system, it could be beneficial if $\lambda$ could be tuned automatically most of the time, and only in critical situations or whenever the user desires they would take over. We thus propose strategies how to find well performing $\lambda$ values at runtime.\\

\textbf{Metrics} - before we introduce the strategies to find $\lambda$ at runtime, we define metrics by which to evaluate them. Depending on the concrete system, different criteria may be of concern: In safety critical systems for example the metric has to reflect that the most important factor is to not enter potentially dangerous system states, while in other systems it is most important to find a good value as quickly as possible, since evaluations incur large opportunity cost as the system cannot be productively used. In many other systems it is likely most important to underperform the behavioral policy as little and as few times as possible in order to facilitate trust in the method and to enable continuous productive use. Finally it could be most important to actually find the single best $\lambda$ value, since the policy will run for a very long time and any underperformance in the search phase is easily offset by performing better in the long run.\\
We consider the metrics Final Return (R), as well as Return under Budget (RUB) indicating the best return possible with a limited budget, Mean Behavioral Regret (MBR) representing the average underperformance during the search phase with respect to the behavioral, and Mean Optimal Regret (MOR) which is the average underperformance in the search phase compared to the best $\lambda$ value. \\

\textbf{Search} - we compare different search strategies to build AutoLION, that aim to optimize different performance metrics. As a baseline, we include the example strategy proposed in \cite{lion}: $\lambda$ is initialized with zero and increased in small steps - as soon as a new return is worse than the previous one, the search stops and the previous $\lambda$ is chosen. We refer to this strategy as Increase-Conservative (Inc-Con). Since this strategy can be considered quite restrictive, we consider a relaxation Increase-Behavioral (Inc-Beh), which stops only once the performance drops below that of the behavioral policy, and then uses the best so far observed $\lambda$ value. These two strategies obviously aim to reduce regret compared to the behavioral performance, which is often a key metric, however we also evaluate a Greedy strategy, which starts at the opposite end and moves towards lower $\lambda$ values in the hope of finding the optimum quicker if the policy can generalize well. Then we compare different gradient-free optimizing strategies that have previously been proposed to see if they can outperform the simple strategies in speed, performance, as well as regret measures: PSO (Particle Swarm Optimization \cite{pso}) is a swarm based optimization method which uses neighbourhood information for search. Scr (ScrHammersleySearchPlusMiddlePoint \cite{nevergrad}) is a one-shot optimizer which hopefully reduces the number of evaluations necessary. DE (Differential Evolution \cite{de}) iteratively improves the candidate solution(s), similar to PSO, and NGOpt \cite{nevergrad} is a meta optimizer that, like Scr, should produce decent solutions even with low evaluation budgets.\\

\section{Experiments}
We use the industrial benchmark (IB) datasets and MuJoCo Swimmer, Hopper, and Walker datasets as proposed in \cite{moose}. The IB is a benchmark motivated by commonly encountered control problems in industrial real-world scenarios, exhibiting transitions with heteroscedastic noise, delayed reward components, partially observable states, and high dimensional state and action spaces. The respective datasets have been gathered by combining three deterministic baseline policies (bad, mediocre, optimized) with different levels of $\varepsilon$-greedy exploration (0\%, 20\%, 40\%, 60\%, 80\%, 100\%). The considered MuJoCo tasks are also well known RL benchmarks with highly complex, yet deterministic and fully observable state transitions. The respective datasets have been gathered by adding different noise sources to an expert policy. Combined, we evaluate AutoLION on 19 datasets from different tasks and domains and a wide variety of settings, which should give a general overview on how the method performs.\\

\begin{figure}[h]
    \centering
    \includegraphics[width=\textwidth]{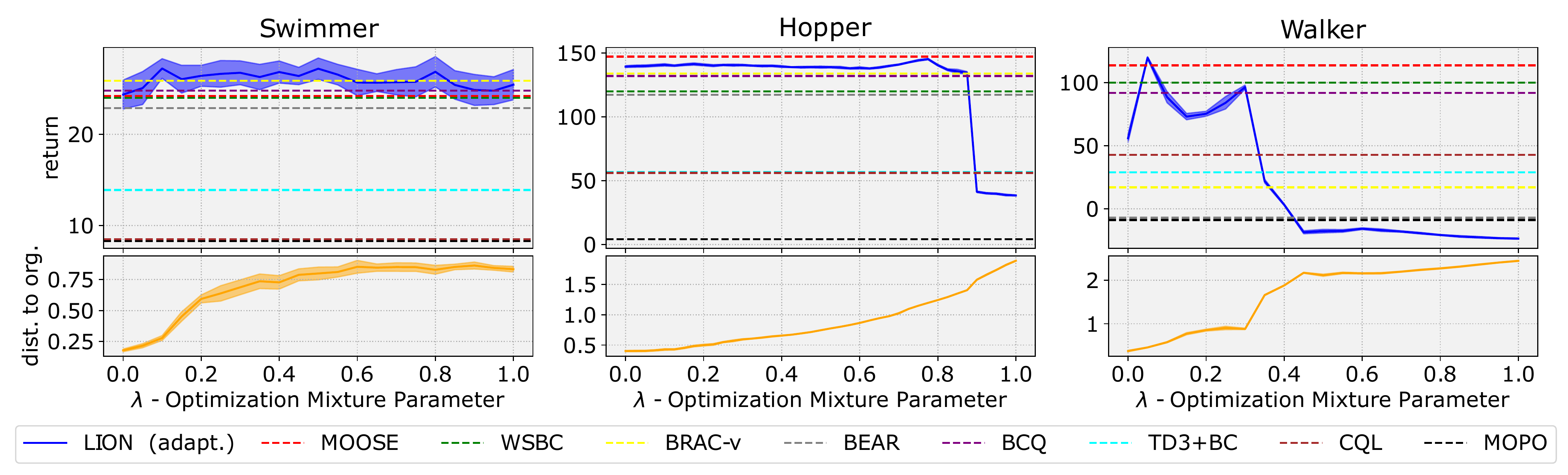}
    \caption{Visualization of the performance landscape of the LION algorithm over the the $\lambda$ spectrum (blue), together with the corresponding proximity to the behavioral policy (orange). We show the same set of offline RL baselines as in \cite{lion} - they are presented as horizontal dashed lines since they cannot adapt the trade-off and since we cannot assign their regularization parameter to a corresponding $\lambda$ value.}
    \label{fig:lion_mujoco}
\end{figure}

\begin{figure}[h]
    \centering
    \begin{minipage}[b]{0.49\textwidth}
    \includegraphics[width=\textwidth]{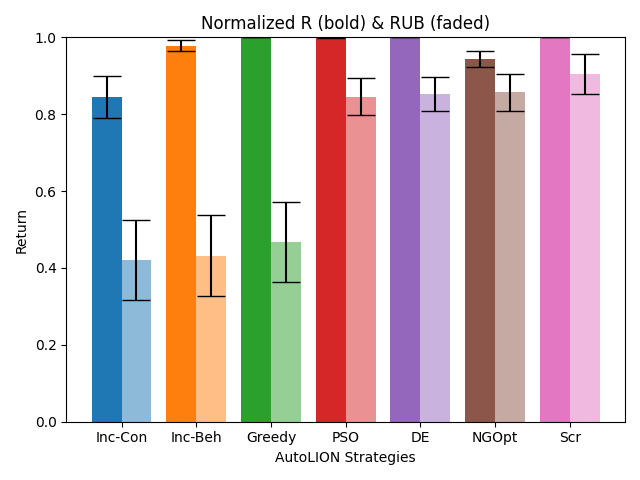}
  \end{minipage}
  \hfill
  \begin{minipage}[b]{0.49\textwidth}
    \includegraphics[width=\textwidth]{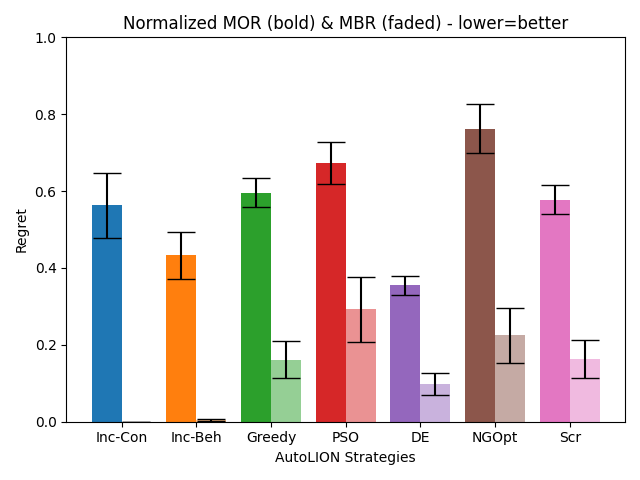}
  \end{minipage}
    \caption{(a) Mean and standard error of normalized unconstrained returns versus returns under budget for each AutoLION strategy, aggregated across all 19 datasets. While Inc-Beh and Greedy are competitive in the unconstrained budget setting, their performance drops significantly when the evaluation budget is limited. The gradient-free search algorithms all appear much more robust in this regard, especially Scr does not loose much performance under the limited budget. (b) Mean and standard error of Normalized Mean Regret w.r.t. the optimal $\lambda$ choice (MOR) and Mean Regret w.r.t. the behavioral choice (i.e. $\lambda=0$). Inc-Con and Inc-Beh are unmatched in their MBR, while Differential Evolution achieves the best MOR (and the best MBR among the gradient-free search strategies). We normalize returns and regrets to the (0,1) interval by subtracting dataset-wise minimum and dividing by (maximum - minimum).}
    \label{fig:eval}
\end{figure}

We train LION policies as proposed in \cite{lion}, and then search the space of trade-off parameters $\lambda$ with the proposed search methods from the previous section in order to provide an automated adaptive offline RL algorithm that works in continuous action spaces. Return and distance to behavior policy curves over the $\lambda$ spectrum for the MuJoCo datasets are provided in Fig. \ref{fig:lion_mujoco}. The evaluations in terms of R, MOR, MBR, as well as RUB are provided in Fig. \ref{fig:eval}: We can see that, for obvious reasons, the Inc-* strategies perform best in terms of regret compared with the behavioral policy, since they use underperformance with respect to the behavioral policy as their stopping criterion (the conservative strategy is even stricter, i.e. it uses an upper bound on the behavioral performance since the return is never allowed to decrease). The conservative strategy has to pay for this with relatively low final returns and is outperformed by any other strategy in our comparison. Our proposed Inc-Beh strategy can be seen as an almost dominant solution compared with the conservative one, since it has close to zero MBR like Inc-Con, but mostly better MOR, and very good R, almost reaching the top performances of Greedy, PSO, Scr, and DE. The Greedy strategy appears competitive with the other gradient-free optimization algorithms in terms of return (especially on the bad datasets), however it is dominated by Scr and DE in every metric, especially in the two regret measurements. match or outperform it while also exhibiting lower MBR. PSO \& NGOpt appear less favourable - they underperform Scr \& DE in terms of regret while achieving similar returns.

\section{Discussion \& Conclusion}
In this paper, we propose to augment conditional, adaptive offline RL policies by a trade-off search phase, which can automatically adjust the corresponding hyperparameter in order to optimize returns once the policy is deployed without any retraining. We introduce relevant performance metrics by which strategies can be evaluated and test the approach with seven different strategies, by augmenting the LION algorithm with our approach, yielding an extension to automatically adapt the trade-off for offline RL at runtime, providing the user an autopilot like option. We thus term our developed approach AutoLION. By testing the algorithm on nineteen datasets from the industrial benchmark and MuJoCo domains, we find that depending on the concrete practical setting, two options dominate: If MBR is critical, i.e. the policy should rarely underperform the behavioral, our proposed relaxation Inc-Beh is likely the best option, while if quick trade-off search under an evaluation budget is needed, Scr performs best. At the intersection between the two, Differential Evolution can represent a meaningful compromise, unifying relatively high returns in both constrained and unconstrained budget settings with relatively low regrets compared to the other gradient-free optimization algorithms. Future work might come up with combinations of the strategies, by e.g. employing MBR-penalized optimizations with few-shot optimizers like Scr or NGOpt.


\begin{footnotesize}





\end{footnotesize}


\end{document}